# Soft Locality Preserving Map (SLPM) for Facial Expression Recognition

Cigdem Turan[a,*], Kin-Man Lam[a], Xiangjian He[b]

[a] Centre for Signal Processing, Department of Electronic and Information Engineering, The Hong Kong Polytechnic University, Kowloon, Hong Kong
[b] Computer Science, School of Electrical and Data Engineering, University of Technology, Sydney, Australia

*Corresponding author.
E-mail addresses: cigdem.turan@connect.polyu.hk (C. Turan), enkmlam@polyu.edu.hk (K.-M. Lam), xiangjian.he@uts.edu.au (X. He)

**ABSTRACT**
For image recognition, an extensive number of methods have been proposed to overcome the high-dimensionality problem of feature vectors being used. These methods vary from unsupervised to supervised, and from statistics to graph-theory based. In this paper, the most popular and the state-of-the-art methods for dimensionality reduction are firstly reviewed, and then a new and more efficient manifold-learning method, named Soft Locality Preserving Map (SLPM), is presented. Furthermore, feature generation and sample selection are proposed to achieve better manifold learning. SLPM is a graph-based subspace-learning method, with the use of *k*-neighbourhood information and the class information. The key feature of SLPM is that it aims to control the level of spread of the different classes, because the spread of the classes in the underlying manifold is closely connected to the generalizability of the learned subspace. Our proposed manifold-learning method can be applied to various pattern recognition applications, and we evaluate its performances on facial expression recognition. Experiments on databases, such as the Bahcesehir University Multilingual Affective Face Database (BAUM-2), the Extended Cohn-Kanade (CK+) Database, the Japanese Female Facial Expression (JAFFE) Database, and the Taiwanese Facial Expression Image Database (TFEID), show that SLPM can effectively reduce the dimensionality of the feature vectors and enhance the discriminative power of the extracted features for expression recognition. Furthermore, the proposed feature-generation method can improve the generalizability of the underlying manifolds for facial expression recognition.

## 1. Introduction

Dimensionality reduction, which aims to find the distinctive features to represent high-dimensional data in a low-dimensional subspace, is a fundamental problem in classification. Many real-world computer-vision and pattern-recognition applications, e.g. facial expression recognition, are involved with large volumes of high-dimensional data. Principal Component Analysis (PCA) [6, 7] and Linear Discriminant Analysis (LDA) [7, 10] are two notable linear methods for dimensionality reduction. PCA aims to find principal projection vectors, which are those eigenvectors associated with the largest eigenvalues of the covariance matrix of training samples, to project the high-dimensional data to a low-dimensional subspace. Unlike PCA, which is an unsupervised method that considers common features of training samples, LDA employs the Fisher criteria to maximize the between-class scattering and to minimize the within-class scattering. Although LDA is superior to PCA for pattern recognition, it suffers from the small-sample-size (SSS) problem [13] because the number of training samples available is much smaller than the dimension of the feature vectors in most of the real-world applications. To overcome the SSS problem, Li et al. [15] proposed the Maximum Margin Criterion (MMC) method, which utilizes the difference between the within-class and the between-class scatter matrices as the objective function.

In [16], it is shown that intra-class scattering has an important effect when dealing with overfitting in training a model. Unlike the conventional wisdom, too much compactness within each class decreases the generalizability of the manifolds. Since LDA and MMC are too "harsh", they need to be softened. Liu et al. [16] proposed the Soft Discriminant Map (SDM), which tries to control the spread of the different classes. MMC can be considered as a special case of SDM, where the softening parameter $\beta = 1$.

Linear methods, like PCA, LDA and SDM, may fail to find the underlying nonlinear structure of the data under consideration, and they may lose some discriminant information of the manifolds during the linear projection. To overcome this problem, some nonlinear dimensionality reduction techniques have been proposed. In general, the nonlinear dimensionality reduction techniques are divided into two categories: kernel-based and manifold-learning-based approaches. Kernel-based methods, as well as the linear methods mentioned above, only employ the global structure while ignoring the local geometry of the data. However, manifold-learning-based methods can explore the intrinsic geometry of the data. Popular nonlinear manifold-learning methods include ISOMAP [17], LLE [18], and Laplacian Eigenmaps [19], which can be considered as special cases of the general framework for dimensionality reduction named "graph embedding", proposed by Yan et al. [5]. Although these methods can represent the local structure of the data, they suffer from the out-of-sample problem. Locality Preserving Projection (LPP) [1] was proposed as a linear approximation of the nonlinear Laplacian Eigenmaps [19] to overcome the out-of-sample problem. LPP considers the manifold structure via the adjacency graph. The manifold-learning methods presented so far are unsupervised methods, i.e. they do not consider the class information. Several supervised-based methods [2, 20, 21] have been proposed, which utilize the discriminant structure of the manifolds. With the Marginal Fisher Analysis (MFA) [5], which uses the Fisher criterion and constructs two adjacency graphs to represent the within-class and the between-class geometry of the data, several other methods have been proposed with similar ideas, such as Locality-Preserved Maximum Information Projection (LPMIP) [9], Constrained Maximum Variance Mapping (CMVM) [8], and Locality Sensitive Discriminant Analysis (LSDA) [4]. In real-life applications, unlabeled data can exist because of various reasons. To deal with this problem, various semi-supervised learning algorithms have also been proposed [22-24].

In this paper, we propose a new graph-based subspace-learning method to solve the various problems of the existing methods by combining their best components to form a better method. The proposed method, named "Soft Locality Preserving Map (SLPM)" can be outlined as follows:

1. SLPM constructs a within-class graph matrix and a between-class graph matrix using the $k$-nearest neighborhood and the class information to discover the local geometry of the data.
2. To overcome the SSS problem and to decrease the computational cost of computing the inverse of a matrix, SLPM defines its objective function as the difference between the between-class and the within-class Laplacian matrices.
3. Inspired by the idea of SDM on the importance of the intra-class spread, a parameter $\beta$ is added to control the penalty on the within-class Laplacian matrix so as to avoid the overfitting problem and to increase the generalizability of the underlying manifold.

Although subspace-learning methods have demonstrated promising performances by increasing the discriminative power of training data after transformation, they might fail to exhibit a similar performance on testing data. To improve the generalizability of the manifolds generated by the subspace-analysis methods, more training samples, which are located near the boundaries of the respective classes, are desirable. In this paper, we apply our proposed SLPM method to facial expression recognition, and propose an efficient way to enhance the generalizability of the manifolds of the different expression classes by feature generation. An expression video sequence, which ranges from a neutral-expression face to the highest intensity of an expression, allows us to select appropriate samples for learning a better and more representative manifold for the expression class. For the optimal manifold of an expression class, its center should represent those samples that best represent the facial expression concerned, i.e. those expression face images with the highest intensities. When moving away from the manifold center, the corresponding expression intensity should be reducing. Those samples near the boundary of a manifold are important for describing the expression, which also defines the shape of the manifold. To describe a manifold boundary, images with low-intensity expressions should be considered. Since the feature vectors used to represent facial expressions usually have high dimensionality, many training samples near the manifold boundary are required, so as to represent it completely. However, we usually have a limited number of weak-intensity expression images, so feature generation is necessary to learn more complete manifolds.

In other applications, additional samples have also been generated for manifold learning. In [25], faces are morphed between two people with different percentages so as to generate face images near the manifold boundaries. By generating more face images and extracting their feature vectors, the manifold for each face subject can be learned more accurately. Therefore, the decision region for each subject can be determined for watch-list surveillance. In our algorithm, rather than morphing faces and extracting features from the synthesized face, we generate features for low-intensity expressions directly in the feature domain. Generating features in this way should be more accurate than extracting features from distorted faces generated by morphing. Several fields of research, such as text categorization [26], handwritten digit recognition [27], facial expression recognition [28, 29], etc., have also employed feature generation to achieve better learning. Unlike these methods, which generate features in the image domain, the proposed method generates features in the feature domain.

In Section 2, we further explain the graph-embedding techniques, and give a detailed comparison of those existing subspace-learning approaches similar to our proposed method. In Section 3, the proposed method, SLPM, is formulated and its relation to SDM is further explored. In Section 4, we explain the local descriptors used in our experiments and the feature-generation algorithm, and describe how to enhance the manifold learning with low-intensity images. In Section 5, we present the databases used in our experiments, and the preprocessing of the face images. Then, experiment results are represented, with a discussion. We conclude this paper in Section 6.

## 2. Literature review of subspace learning

In this section, a review of the graph-embedding techniques is presented in detail, with the different variants. Then, graph-based subspace-learning methods are described in two parts: 1) how the adjacency matrices are constructed, and 2) how their objective functions are defined.

### 2.1. Graph embedding

Given $m$ data points $\{x_1, x_2, \ldots, x_m\} \in \mathbb{R}^D$, the graph-based subspace-learning methods aim to find a transformation matrix $\mathbf{A}$ that maps the training data points to a new set of points $\{y_1, y_2, \ldots, y_m\} \in \mathbb{R}^d$ ($d \ll D$), where $y_i = \mathbf{A}^T x_i$ and $\mathbf{A}$ is the projection matrix. After the transformation, the data points $x_i$ and $x_j$, which are close to each other, will have their projections in the manifold space $y_i$ and $y_j$ close to each other. This goal can be achieved by minimizing the following objective function:

$$\sum_{ij}(y_i - y_j)^2 w_{ij}, \qquad (1)$$

where $w_{ij}$ represents the similarity between the training data $x_i$ and $x_j$. If $w_{ij}$ is non-zero, $y_i$ and $y_j$ must be close to each other, in order to minimize (1). Taking the data points in the feature space as nodes of a graph, an edge between nodes $i$ and $j$ has a weight of $w_{ij}$, which is not zero, if they are close to each other. In the literature, we have found three different ways to determine the local geometry of a data point:

1. **ε-neighbourhood:** This uses the distance to determine the closeness. Given ε (ε ∈ ℝ), ε-neighbourhood chooses the data points that fall within the circle around $x_i$ with a radius ε. Those data points fall within the ε-neighbourhood of $x_i$ can be defined as

$$O(x_i, \varepsilon) = \{x \mid \|x - x_i\|^2 < \varepsilon\}. \qquad (2)$$

2. **$k$-nearest neighbourhood:** Another way of determining the local structure is to use the nearest neighbourhood information. Presuming that the closest $k$ points of $x_i$ would still be the closest data points of $\mathbf{y}_i$ in the projected manifold space, we can define a function $N(\mathbf{x}_i, k)$, which outputs the set of $k$-nearest neighbours of $\mathbf{x}_i$. Two types of neighbourhood, with label information incorporated, are considered: $N(\mathbf{x}_i, k^+)$ and $N(\mathbf{x}_i, k^-)$, which represent the sets of $k$-nearest neighbours of $\mathbf{x}_i$ of the same label and of different labels, respectively.

3. **The class information:** The class or label information is often used in supervised subspace methods. In a desired manifold subspace, the data points belonging to the class of $\mathbf{x}_i$ are to be projected such that they are close to each other, so as to increase the intra-class compactness. The data points belonging to other classes are projected, such that they will become farther apart and have larger inter-class separability. The class label information is often combined with either the ε-neighbourhood or the $k$-nearest neighbourhood.

The similarity graph is constructed by setting up edges between the nodes. There are different ways of determining the weights of the edges, considering the fact that the distance between two neighbouring points can also provide useful information about the manifold. Given a sparse symmetric similarity matrix $\mathbf{W}$, two variations have been proposed in the literature.

1. Binary weights: $w_{ij} = 1$ if, and only if, the nodes $i$ and $j$ are connected by an edge, otherwise $w_{ij} = 0$.
2. Heat kernel ($t \in \mathbb{R}$): If the nodes $i$ and $j$ are connected by an edge, the weight of the edge is defined as
$$w_{ij} = \exp\left(-\|\mathbf{x}_i - \mathbf{x}_j\|^2 / t\right). \tag{3}$$

After constructing the similarity matrix with the weights, the minimization problem defined in (1) can be solved by using the spectral graph theory. Defining the Laplacian matrix $\mathbf{L} = \mathbf{D} - \mathbf{W}$, where $\mathbf{D}$ is the diagonal matrix whose entries are the column sum of $\mathbf{W}$, i.e. $d_{ii} = \sum_j w_{ij}$, the objective function is reduced to

$$\min \sum_{ij}(\mathbf{y}_i - \mathbf{y}_j)^2 w_{ij} = \min \sum_{ij}(\mathbf{A}^\mathrm{T}\mathbf{x}_i - \mathbf{A}^\mathrm{T}\mathbf{x}_j)^2 w_{ij}$$
$$= \min \mathbf{A}^\mathrm{T}\mathbf{XLX}^\mathrm{T}\mathbf{A}, \tag{4}$$

where $\mathbf{X} = [\mathbf{x}_1, \mathbf{x}_2, \dots, \mathbf{x}_m]$. To avoid the trivial solution of the objective function, the constraint $\mathbf{A}^\mathrm{T}\mathbf{XDX}^\mathrm{T}\mathbf{A} = 1$ is often added. After specifying the objective function, the optimal projection matrix $\mathbf{A}$ can be obtained by choosing the eigenvectors corresponding to the $d$ ($d \ll D$) smallest non-zero eigenvalues computed by solving the standard eigenvalue decomposition or generalized eigenvalue problem, depending on the objective function being considered.

2.1.1. *Constructing the within-class and the between-class graph matrices*

As mentioned in the last section, one of the most popular graph-based subspace-learning methods is LPP [1], which uses an intrinsic graph to represent the locality information of the dataset, i.e. the neighbourhood information. The idea behind LPP is that if the data points $\mathbf{x}_i$ and $\mathbf{x}_j$ are close to each other in the feature space, then they should also be close to each other in the manifold subspace. The similarity matrix $w_{ij}$ for LPP can be defined as follows:

$$w_{ij} = \begin{cases} 1, & \text{if } \mathbf{x}_i \in N(\mathbf{x}_j, k) \text{ or } \mathbf{x}_j \in N(\mathbf{x}_i, k), \\ 0, & \text{otherwise}, \end{cases} \tag{5}$$

where $N(\mathbf{x}_i, k)$ represents the set of $k$-nearest neighbors of $\mathbf{x}_i$. One shortfall of the above formulation for $w_{ij}$ is that it is an unsupervised method, i.e. not using any class-label information. Thinking that the label information can help to find a better separation between different class manifolds, Supervised Locality Preserving Projections (SLPP) was introduced in [2]. Denote $l(\mathbf{x}_i)$ as the corresponding class label of the data point $\mathbf{x}_i$. SLPP uses either one of the following formulations:

$$w_{ij} = \begin{cases} 1, & \text{if } l(\mathbf{x}_i) = l(\mathbf{x}_j), \\ 0, & \text{otherwise}, \end{cases} \tag{6}$$

$$w_{ij} = \begin{cases} 1, & \text{if } (x_i \in N(x_j, k) \text{ or } x_j \in N(x_i, k)) \text{ and } l(x_i) = l(x_j), \\ 0, & \text{otherwise.} \end{cases} \quad (7)$$

Note that (6) does not include the neighbourhood information to the adjacency graph, and the similarity matrices defined above can be constructed using the heat kernel. Orthogonal Locality Preserving Projection (OLPP) [3] whose eigenvectors are orthogonal to each other is an extension of LPP. Please note that, in our experiments we applied Supervised Orthogonal Locality Preserving Projections (SOLPP), which is OLPP with its adjacency matrix including class information.

Yan et al. [5] proposed a general framework for dimensionality reduction, named Marginal Fisher Analysis (MFA). MFA, which is based on graph embedding as LPP, uses two graphs, the intrinsic and penalty graphs, to characterize the intra-class compactness and the interclass separability, respectively. In MFA, the intrinsic graph $w_{ij}^w$, i.e. the within-class graph, is constructed using the neighbourhood and class information as follows:

$$w_{ij}^w = \begin{cases} 1, & \text{if } x_i \in N(x_j, k_1^+) \text{ or } x_j \in N(x_i, k_1^+), \\ 0, & \text{otherwise.} \end{cases} \quad (8)$$

Similarly, the penalty graph $w_{ij}^b$, i.e. the between-class graph, is constructed as follows:

$$w_{ij}^b = \begin{cases} 1, & \text{if } x_i \in N(x_j, k_2^-) \text{ or } x_j \in N(x_i, k_2^-), \\ 0, & \text{otherwise.} \end{cases} \quad (9)$$

Locality Sensitive Discriminant Analysis (LSDA) [4] and Improved Locality Sensitive Discriminant Analysis (ILSDA) [14] are subspace-learning methods proposed in 2007 and 2015, respectively. They construct the similarity matrices in the same way, but LSDA uses binary weights, while ILSDA sets the weight of the edges using the heat kernel. The similarity matrices of LSDA are defined as follows:

$$w_{ij}^w = \begin{cases} 1, & \text{if } x_i \in N(x_j, k) \text{ and } l(x_i) = l(x_j), \\ 0, & \text{otherwise.} \end{cases} \quad (10)$$

$$w_{ij}^b = \begin{cases} 1, & \text{if } x_i \in N(x_j, k) \text{ and } l(x_i) \neq l(x_j), \\ 0, & \text{otherwise.} \end{cases} \quad (11)$$

It can be observed that the intrinsic and the penalty graphs of MFA, LSDA, and ILSDA are similar to each other. In MFA, the numbers of neighboring points for both the similarity matrices are known, i.e. $k_1$ and $k_2$. In LSDA and ILSDA, the $k$ neighbors of $x_i$ are selected, which are then divided for constructing the within-class ($k^+$ samples the same class as $x_i$) and the between-class matrices ($k^-$ samples of other classes), i.e. $k = k^+ + k^-$. Let $k_1$ and $k_2$ be the numbers of samples belonging to the same class and different classes, respectively, for MFA. The following relation is not always true:

$$N(x_i, k_1) \cup N(x_i, k_2) = N(x_i, k), \quad (12)$$

because it is not necessarily true that $k^+ = k_1$ and $k^- = k_2$. Therefore, the neighboring points of $x_i$ in LSDA and ILSDA are not the same as MFA, even if $k = k_t = k_1 + k_2$. However, the adjacency matrices constructed in the manifold learning methods are similar to each other. The main difference between the existing methods in the literature is in their definitions of the objective functions. We will elaborate on the differences in the objective functions in the next section.

Locality-Preserved Maximum Information Projection (LPMIP) [9], proposed in 2008, uses the ε-neighbourhood condition, i.e. $O(x_i, \varepsilon)$. Although it was originally applied as an unsupervised learning method, the class labels were used to construct the locality and non-locality information for facial expression recognition. In 2008, Li et al. [8] proposed Constrained Maximum Variance Mapping (CMVM), which aims to keep the local structure of the data, while separating the different manifolds, i.e. different classes, farther apart. The local-structure graphs, i.e. the between-class graph and the dissimilarities graph, are defined as follows:

$$w_{ij}^w = \begin{cases} 1 \text{ or } \exp\left(-\|x_i - x_j\|^2/t\right), & \text{if } x_i \in O(x_j, \varepsilon), \\ 0, & \text{otherwise,} \end{cases} \quad (13)$$

$$w_{ij}^b = \begin{cases} 1, & \text{if } l(x_i) \neq l(x_j), \\ 0, & \text{otherwise.} \end{cases} \tag{14}$$

As (13) and (14) show, the within-class matrix of CMVM only preserves the local structure of the whole data, while the between-class matrix only uses the class label to increase the separability of different class manifolds. In 2015, an extension of CMVM, namely CMVM+ [12], was proposed to overcome the obstacles of CMVM. CMVM+ adds the class information and neighbourhood information to the similarity matrices. The updated version of the graphs can be written as follows:

$$w_{ij}^w = \begin{cases} 1, & x_i \in N(x_j, k) \text{ and } l(x_i) = l(x_j), \\ 0, & \text{otherwise,} \end{cases} \tag{15}$$

$$w_{ij}^b = \begin{cases} 1, & \text{if } l(x_j) \in C_{inc}(x_i), \\ 0, & \text{otherwise,} \end{cases} \tag{16}$$

where $C_{inc}(x_i)$ is a set of neighboring points belonging to different classes, i.e. $l(x_i) \neq l(x_j)$. More details of the function $C_{inc}(x_i)$ can be found in [12].

In 2011, Multi-Manifolds Discriminant Analysis (MMDA) [11] was proposed for image feature extraction, and applied to face recognition. The idea behind MMDA is to keep the points from the same class as close as possible in the manifold space, with the within-class matrix defined as follows:

$$w_{ij}^w = \begin{cases} \exp\left(-\|x_i - x_j\|^2 / t\right), & \text{if } l(x_i) = l(x_j), \\ 0, & \text{otherwise.} \end{cases} \tag{17}$$

MMDA also constructs a between-class matrix in order to separate the different classes from each other. The difference between the between-class matrix of MMDA and the other subspace methods is that its graph matrix is constructed by not taking all the data points as nodes, but rather calculating the weighted centres of different classes by averaging all the data points belonging to the classes under consideration. Let $\mathbf{M} = [\widetilde{m}_1, \widetilde{m}_2, \dots, \widetilde{m}_c]$ be the class-weighted centres, where $c$ is the number of classes. Then, the between-class matrix of MMDA can be written as:

$$w_{ij}^b = \exp\left(-\|\widetilde{m}_i - \widetilde{m}_j\|^2 / t\right). \tag{18}$$

In Table 1, a summary is given of the within-class graph and between-class graph for the subspace-learning methods, reviewed in this paper.

### 2.1.2. *Defining the objective functions*

Table 2 summarizes the objective functions of the approaches reviewed in the previous section, as well as the constraints used. We can see that SLPP has only one Laplacian matrix defined in its objective function, because it constructs one similarity matrix only, while all the other methods have two matrices: one is based on the intrinsic graph, and the other on the penalty graph.

In general, there are two ways of defining the objective functions with the intrinsic and the penalty matrices. The first one utilizes the Fisher criterion to maximize the ratio between the scattering of the between-class and that of the within-class Laplacian matrices. MFA, MMDA, and CMVM+ employ the Fisher criterion. Although the application of the Fisher criterion shows its robustness, it involves taking the inverse of a high-dimensional matrix to solve a generalized eigenvalue problem. To solve this problem, LSDA, LPMIP, and our proposed method define the objective functions as the difference between the intrinsic and the penalty-graph matrices, while MMC and SDM use the difference between the inter-class and the intra-class scatter matrices.

As shown in Table 2, ILSDA adopts a similar objective function to LSDA, but with a difference that the within-class scatter matrix is included in the objective function. The within-class scatter matrix $\mathbf{S}_w$ — as used in LDA — indicates the compactness of the data point in each class. ILSDA uses the scatter matrix to project outliers closer to the class centers under consideration. The objective function of ILSDA is defined as follows:

$$\max \mathbf{A}^T(\mathbf{P} - \alpha \mathbf{S}_w)\mathbf{A}, \quad (19)$$

where $\mathbf{P} = \mathbf{X}(\mathbf{L}_b - \mathbf{L}_w)\mathbf{X}^T$, as defined in the objective function of LSDA. CMVM, unlike other methods which aim to minimize the within-class spread, intends to maintain the within-class structure for each class by defining a constraint, i.e. $\mathbf{A}^T\mathbf{X}\mathbf{L}_w\mathbf{X}^T\mathbf{A} = \mathbf{X}\mathbf{L}_w\mathbf{X}^T$, while increasing the inter-class separability with the following objective function:

$$\max \mathbf{A}^T\mathbf{X}\mathbf{L}_b\mathbf{X}^T\mathbf{A}, \quad (20)$$

where $\mathbf{L}_w$ and $\mathbf{L}_b$ are the within-class and the between-class Laplacian matrices, respectively.

Table 1. A comparison of the within-class graph and the between-class graph for different subspace-learning methods. (bn: binary weights, hk: heat kernel, knn: *k*-nearest neighborhood)

| Subspace Learning Methods | The within-class graph | | | The between-class graph | | |
|---|---|---|---|---|---|---|
| | Neighbourhood | Class Info | Weight | Neighbourhood | Class Info | Weight |
| LPP [1]/ OLPP [3] | optional | No | optional | n/a | n/a | n/a |
| SLPP [2]/ SOLPP | optional | Yes | optional | n/a | n/a | n/a |
| LSDA [4] | knn | Yes | bn | knn | Yes | bn |
| MFA [5] | knn | Yes | bn | knn | Yes | bn |
| CMVM [8] | ϵ-ball | No | bn/hk | n/a | Yes | bn |
| LPMIP [9] | ϵ-ball | No | hk | ϵ-ball | No | hk |
| MMDA [11] | n/a | Yes | hk | class centers | Yes | hk |
| CMVM+ [12] | knn | Yes | bn | knn | Yes | bn |
| ILSDA [14] | knn | Yes | hk | knn | Yes | hk |
| SLPM (proposed) | knn | Yes | bn/hk | knn | Yes | hk |

## 3. Soft locality preserving map

In this section, we introduce the proposed method, Soft Locality Preserving Map (SLPM), with its formulation and connection to the previous works. Then, we will also describe the local descriptors used for facial expression recognition in our experiments.

### 3.1. Formulation of the SLPM

Similar to other manifold-learning algorithms, two graph-matrices, the between-class matrix $\mathbf{W}_b$ and the within-class matrix $\mathbf{W}_w$, are constructed to characterize the discriminative information, based on the locality and class-label information. Given $m$ data points $\{\mathbf{x}_1, \mathbf{x}_2, \ldots, \mathbf{x}_m\} \in \mathbb{R}^D$ and their corresponding class labels $\{l(\mathbf{x}_1), l(\mathbf{x}_2), \ldots, l(\mathbf{x}_m)\}$, we denote $N_w(\mathbf{x}_i, k_w) = \{\mathbf{x}_i^{w_1}, \mathbf{x}_i^{w_2}, \ldots, \mathbf{x}_i^{w_{k_w}}\}$ as the set of $k_w$ nearest neighbours with the same class label as $\mathbf{x}_i$, i.e. $l(\mathbf{x}_i) = l(\mathbf{x}_i^{w_1}) = l(\mathbf{x}_i^{w_2}) = \cdots = l(\mathbf{x}_i^{w_{k_w}})$, and $N_b(\mathbf{x}_i, k_b) = \{\mathbf{x}_i^{b_1}, \mathbf{x}_i^{b_2}, \ldots, \mathbf{x}_i^{b_{k_b}}\}$ as the set of its $k_b$ nearest neighbours with different class labels from $\mathbf{x}_i$, i.e. $l(\mathbf{x}_i) \neq l(\mathbf{x}_i^{w_j})$, where $j = 1, 2, \ldots, k_b$. Then, the inter-class weight matrix $\mathbf{W}_b$ and the intra-class weight matrix $\mathbf{W}_w$ can be defined as below:

Table 2. A comparison of the objective functions used by different subspace methods.

| | **Objective functions** | **Constraints (s.t.)** |
|---|---|---|
| **LPP [1]/ SLPP [2]** | $\max_{\mathbf{A}} \mathbf{A}^T \mathbf{XLX}^T \mathbf{A}$ | $\mathbf{A}^T \mathbf{XDX}^T \mathbf{A} = \mathbf{I}$ |
| **LSDA [4]** | $\max_{\mathbf{A}} \mathbf{A}^T \mathbf{X}(a\mathbf{L}_b + (1-\alpha)\mathbf{W}_w)\mathbf{X}^T \mathbf{A}$ | $\mathbf{A}^T \mathbf{XD}_w \mathbf{X}^T \mathbf{A} = \mathbf{I}$ |
| **MFA [5]** | $\min_{\mathbf{A}} \dfrac{\mathbf{A}^T \mathbf{XL}_w \mathbf{X}^T \mathbf{A}}{\mathbf{A}^T \mathbf{XL}_b \mathbf{X}^T \mathbf{A}}$ | n/a |
| **CMVM [8]** | $\max \mathbf{A}^T \mathbf{XL}_b \mathbf{X}^T \mathbf{A}$ | $\mathbf{A}^T \mathbf{XL}_w \mathbf{X}^T \mathbf{A} = \mathbf{XL}_w \mathbf{X}^T$ |
| **LPMIP [9]** | $\max_{\mathbf{A}} \mathbf{A}^T \mathbf{X}(a\mathbf{L}_b - (1-\alpha)\mathbf{L}_w)\mathbf{X}^T \mathbf{A}$ | $\mathbf{A}^T \mathbf{A} - \mathbf{I} = 0$ |
| **MMDA [11]** | $\max_{\mathbf{A}} \dfrac{\mathbf{A}^T \mathbf{XL}_b \mathbf{X}^T \mathbf{A}}{\mathbf{A}^T \mathbf{XL}_w \mathbf{X}^T \mathbf{A}}$ | n/a |
| **CMVM+ [12]** | $\max_{\mathbf{A}} \dfrac{\mathbf{A}^T \mathbf{XL}_b \mathbf{X}^T \mathbf{A}}{\mathbf{A}^T \mathbf{XL}_w \mathbf{X}^T \mathbf{A}}$ | n/a |
| **ILSDA [14]** | $\max \mathbf{A}^T (\mathbf{P} - \alpha \mathbf{S}_w) \mathbf{A}$ where $\mathbf{P} = \mathbf{X}(\mathbf{L}_b - \mathbf{L}_w)\mathbf{X}^T$ | $\mathbf{A}^T \mathbf{A} = \mathbf{I}$ |
| **SDM [16]** | $\max \mathbf{S}_b - \alpha \mathbf{S}_w$ | n/a |
| **SLPM** | $\max \mathbf{A}^T (\mathbf{XL}_b \mathbf{X}^T - \beta \mathbf{XL}_w \mathbf{X}^T) \mathbf{A}$ <u>or</u> $\max \mathbf{A}^T \mathbf{X}(\mathbf{L}_b - \beta \mathbf{L}_w) \mathbf{X}^T \mathbf{A}$ | $\mathbf{A}^T \mathbf{A} - \mathbf{I} = 0$ |

$$w_{ij}^b = \begin{cases} \exp\left(-\|\mathbf{x}_i - \mathbf{x}_j\|^2 / t\right), & \mathbf{x}_j \in N_b(\mathbf{x}_i, k_b), \\ 0, & \text{otherwise.} \end{cases} \quad (21)$$

$$w_{ij}^w = \begin{cases} \exp\left(-\|\mathbf{x}_i - \mathbf{x}_j\|^2 / t\right), & \mathbf{x}_j \in N_w(\mathbf{x}_i, k_w), \\ 0, & \text{otherwise.} \end{cases} \quad (22)$$

SLPM is a supervised manifold-learning algorithm, which aims to maximize the between-class separability, while controlling the within-class spread with a control parameter $\beta$ used in the objective function. Consider the problem of creating a subspace, such that data points from different classes, i.e. represented as edges in $\mathbf{W}_b$, stay as distant as possible, while data points from the same class, i.e. represented as edges in $\mathbf{W}_w$, stay close to each other. To achieve this, two objective functions are defined as follows:

$$\max \tfrac{1}{2} \sum_{ij} (\mathbf{y}_i - \mathbf{y}_j)^2 w_{ij}^b, \quad (23)$$

$$\min \tfrac{1}{2} \sum_{ij} (\mathbf{y}_i - \mathbf{y}_j)^2 w_{ij}^w. \quad (24)$$

Eq. (23) ensures that the samples from different classes will stay as far as possible from each other, while Eq. (24) is to make samples from the same class stay close to each other after the projection.

However, as shown in [30] and [16], small variations in the manifold subspace can lead to overfitting in training. To overcome this problem, we add the parameter $\beta$ to control the intra-class spread. Note that, the method SDM in [16] uses the within-class scatter matrix $\mathbf{S}_w$ – as defined for LDA – to control the intra-class spread. In our proposed method, we adopt the graph-embedding method, which uses the locality information about each class, in addition to the class information. Hence, the two objective functions Eq. (23) and Eq. (24) can be combined as follows:

$$\max \frac{1}{2}\left(\sum_{ij}(\mathbf{y}_i - \mathbf{y}_j)^2 w_{ij}^b - \beta \sum_{ij}(\mathbf{y}_i - \mathbf{y}_j)^2 w_{ij}^w\right)$$
$$= \max(J_b(\mathbf{A}) - \beta J_w(\mathbf{A})), \qquad (25)$$

where $\mathbf{A}$ is a projection matrix, i.e. $\mathbf{Y} = \mathbf{A}^T\mathbf{X}$ and $\mathbf{X} = [\mathbf{x}_1, \mathbf{x}_2, \dots, \mathbf{x}_m]$. Then, the between-class objective function $J_b(\mathbf{A})$ can be reduced to

$$J_b(\mathbf{A}) = \frac{1}{2}\sum_{ij}(\mathbf{y}_i - \mathbf{y}_j)^2 w_{ij}^b$$
$$= \mathbf{A}^T\mathbf{X}\mathbf{L}_b\mathbf{X}^T\mathbf{A} \qquad (26)$$

where $\mathbf{L}_b = \mathbf{D}_b - \mathbf{W}_b$ is the Laplacian matrix of $\mathbf{W}_b$ and $d_{b\,ii} = \sum_j w_{ij}^b$ is a diagonal matrix. Similarly, the within-class objective function $J_w(\mathbf{A})$ can be written as

$$J_w(\mathbf{A}) = \frac{1}{2}\sum_{ij}(\mathbf{y}_i - \mathbf{y}_j)^2 w_{ij}^w$$
$$= \mathbf{A}^T\mathbf{X}\mathbf{L}_w\mathbf{X}^T\mathbf{A} \qquad (27)$$

where $\mathbf{L}_w = \mathbf{D}_w - \mathbf{W}_w$ and $d_{w\,ii} = \sum_j w_{ij}^w$. If $J_w$ and $J_b$ are substituted to Eq. (25), the objective function becomes as follows:

$$\max J_T(\mathbf{A}) = \max(J_b(\mathbf{A}) - \beta J_w(\mathbf{A}))$$
$$= \max(\mathbf{A}^T\mathbf{X}\mathbf{L}_b\mathbf{X}^T\mathbf{A} - \beta \mathbf{A}^T\mathbf{X}\mathbf{L}_w\mathbf{X}^T\mathbf{A})$$
$$= \max \mathbf{A}^T\mathbf{X}(\mathbf{L}_b - \beta\mathbf{L}_w)\mathbf{X}^T\mathbf{A} \qquad (28)$$

which is subject to $\mathbf{A}^T\mathbf{A} - 1 = 0$ so as to guarantee orthogonality. By using Lagrange multiplier, we obtain

$$L(\mathbf{A}) = \mathbf{A}^T\mathbf{X}(\mathbf{L}_b - \beta\mathbf{L}_w)\mathbf{X}^T\mathbf{A} - \lambda(\mathbf{A}^T\mathbf{A} - \mathbf{I}). \qquad (29)$$

By computing the partial derivative of L(**A**), the optimal projection matrix **A** can be obtained, as follows:

$$\frac{\partial L(\mathbf{A})}{\partial \mathbf{A}} = \mathbf{X}(\mathbf{L}_b - \beta\mathbf{L}_w)\mathbf{X}^T\mathbf{A} - \lambda\mathbf{A} = 0, \qquad (30)$$

i.e. $\mathbf{X}(\mathbf{L}_b - \beta\mathbf{L}_w)\mathbf{X}^T\mathbf{A} = \lambda\mathbf{A}$. The projection matrix $\mathbf{A}$ can be obtained by computing the eigenvectors of $\mathbf{X}(\mathbf{L}_b - \beta\mathbf{L}_w)\mathbf{X}^T$. The columns of $\mathbf{A}$ are the $d$ leading eigenvectors, where $d$ is the dimension of the subspace. LDA, LPP, MFA, and other manifold-learning algorithms, whose objective functions have a similar structure, lead to a generalized eigenvalue problem. Such methods suffer from the matrix-singularity problem, because the solution involves computing the inverse of a singular matrix. The proposed objective function is designed in such a way as to overcome this singularity problem. However, in our algorithm, PCA is still applied to data, so as to reduce its dimensionality and to reduce noise.

### 3.2. Intra-class spread

As we have mentioned before, the manifold spread of the different classes can affect the generalizability of the learned classifier. To control the spread of the classes, the parameter $\beta$ is adjusted in our proposed method, like SDM. Figure 1 shows the change in the spread of the classes when $\beta$ increases. We can see that increasing $\beta$ will also increase the separability of the data, e.g. the training data is located at almost the same position in the subspace when $\beta = 1,000$.

Figure 1. The spread of the respective expression manifolds when the value of $\beta$ increases from 1 to 1,000: (1) Anger, (2) Disgust, (3) Fear, (4) Happiness, (5) Sadness, and (6) Surprise.

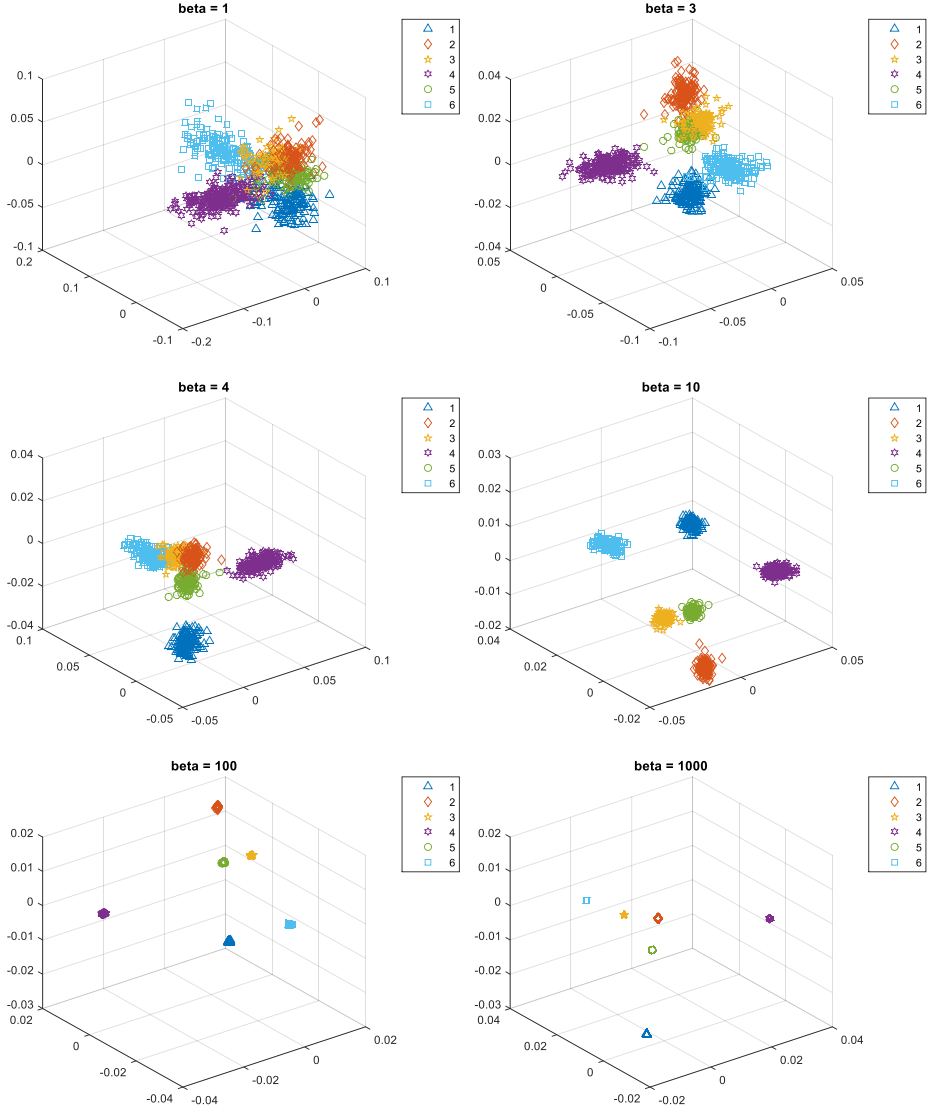

### 3.3. Relations to other subspace-learning methods

As discussed in Section 2, there have been extensive studies on manifold-learning methods. They share the same core idea, i.e. using locality and/or label information to define an objective function, so that the data can be represented in a specific way after projection.

There are two main differences between SLPM and LSDA. First, LSDA defines their objective function as a subtraction of two objective functions like SLPM. However, LSDA imposes the constraint $\mathbf{A}^T\mathbf{X}\mathbf{D}_w\mathbf{X}^T\mathbf{A} = \mathbf{I}$, which results in a generalized eigenvalue problem. As we mentioned in Section 2, the generalized eigenvalue problem suffers from the computational cost of calculating an inverse matrix. SLPM only determines the orthogonal projections, with the constraint $\mathbf{A}^T\mathbf{A} - \mathbf{I} = 0$. Therefore, SLPM can still be computed by eigenvalue decomposition, without requiring computing any

Figure 2. The overall flow of our proposed method.

1. Extract features from face images: $\mathbf{X}_{desc}$
2. Learn the projection matrix $\mathbf{W}_{pca}$ via PCA
3. Construct the within-class graph matrix $\mathbf{W}_w$ and the between-class similarity matrices $\mathbf{W}_b$
4. Calculate the Laplacian matrices $\mathbf{L}_w$ and $\mathbf{L}_b$
5. Solve the eigenvalue decomposition of $\mathbf{X}(\mathbf{L}_b - \beta\mathbf{L}_w)\mathbf{X}^T$
6. Choose the eigenvectors corresponding to the $d$ largest eigenvalues, $\mathbf{W}_{mL}$
7. $\mathbf{Y}_{desc} = \mathbf{W}_{mL}^T \mathbf{W}_{pca}^T \mathbf{X}_{desc}$
8. Add features obtained with either low-intensity images ($\mathbf{Y}^l$) or feature generation ($\overline{\mathbf{Y}}^l$) to form the training data $\mathbf{T}^l$ or $\overline{\mathbf{T}}_l$, respectively
9. Learn the nearest neighbor classifier

inverse matrix. Second, LSDA finds the neighboring points followed by determining whether the considered neighboring points are of the same class or of different classes. This may lead to an unbalanced and unwanted division of neighboring points, simply because of the fact that a sample point may be surrounded by more samples belonging to the same class than samples with different class labels. In order not to lose locality information in such a case, SLPM defines two parameters $k_1$ and $k_2$, which are the numbers of neighboring points belonging to the same class and different classes, respectively. In other words, the numbers of neighboring points belonging to the same class and different classes can be controlled.

Both SDM and ILSDA also consider the intra-class spread when defining the objective function. SDM controls the level of spread by applying a parameter to the within-class scatter matrix $\mathbf{S}_w$. However, it only uses the label information about the training data – its scatter matrices do not consider the local structure of the data. Our proposed SLPM aims to include the locality information by employing graph embedding in our objective functions. Therefore, SLPM is a graph-based version of SDM. ILSDA uses both the label and neighborhood information represented in the adjacent matrices, and also aims to control the spread of the classes. However, ILSDA achieves this by adding the scatter matrix $\mathbf{S}_w$ to its objective function. In our algorithm, we propose controlling the spread with the within-class Laplacian matrix $\mathbf{L}_w$, without adding a separate element to the objective function.

## 4. Feature descriptors and generation

In this section, we will first present the descriptors used for representing facial images for expression recognition, then investigate the use of face images with low-intensity and high-intensity expressions for manifold learning, which represent the corresponding samples at the core and boundary of the manifold for an expression. After that, we will introduce our proposed feature-generation algorithm.

### 4.1. Descriptors

Recent research has shown that local features can achieve higher and more robust recognition performance than by using global features, such as Eigenfaces and Fisherfaces, and intensity values. Therefore, in order to show the robustness of our proposed method, four different commonly used local descriptors for facial expression recognition, Local Binary Pattern (LBP) [31, 32], Local Phase Quantization (LPQ) [33], Pyramid of Histogram of Oriented Gradients (PHOG) [34], and Weber Local Descriptor (WLD) [35], are considered in our experiments. These descriptors can represent face images, in terms of different aspects such as intensity, phase, shape, etc., so that they are complementary to each other. As shown in Figure 2, features are extracted using one of the above-mentioned local descriptors, followed by the subspace learning with SLPM and a feature-generation method.

## 4.2. Feature generation

Features in a projected subspace still have a high dimension. A large number of samples for each expression is necessary in order to represent its corresponding manifold accurately. By generating more features located near the manifold boundaries, more accurate decision boundaries can then be determined for accurate facial expression.

Video sequences with face images, changing from neutral expression to a particular expression, are used for learning. Let $f_{i,\theta}$ denote the frame index of the face image of expression intensity $\theta$ ($0 \leq \theta \leq 1$, $0$ = neutral expression and $1$ = the highest intensity of an expression, i.e. the peak expression) of the sequence $S_i$ in a dataset of $m$ video sequences. Let $x_i^\theta \in \mathbb{R}^D$ be the feature vector extracted from the $f_{i,\theta}$-th frame of the sequence $S_i$. The frame index $f_{i,\theta}$ can be calculated as follows:

$$f_{i,\theta} = n_i \times \theta, \tag{33}$$

where $n_i$ is the number of frames in the sequence $S_i$. Therefore, $\{x_1^1, x_2^1, \ldots, x_m^1\} \in \mathbb{R}^D$ are the feature vectors extracted from the face images with high-intensity expressions, i.e. the last frames of the $m$ video sequences. Suppose that $\{x_1^\xi, x_2^\xi, \ldots, x_m^\xi\}$ are the feature vectors extracted from the corresponding low-intensity images, and the corresponding frame number in the respective video sequences is $f_{i,\xi}$. In our algorithm, we use a different set of $\xi$ values, where $0.6 \leq \xi \leq 0.9$, to learn the different expression manifolds.

### 4.2.1. *Manifold learning with high and low-intensity training samples*

A projection matrix $\mathbf{A}$ that maps the feature vectors $\mathbf{X}^1 = [x_1^1, x_2^1, \ldots, x_m^1]$ to a new subspace is first calculated using SLPM. The corresponding projected samples are denoted as $\mathbf{Y}^1 = [y_1^1, y_2^1, \ldots, y_m^1] \in \mathbb{R}^d$ ($d \ll D$), i.e. $y_i^1 = \mathbf{A}^T x_i^1$. Then, the same projection matrix $\mathbf{A}$ is used to map the low-intensity feature vectors $\mathbf{X}^\xi = [x_1^\xi, x_2^\xi, \ldots, x_m^\xi]$, i.e. $y_i^\xi = \mathbf{A}^T x_i^\xi$, which should lie on the boundary of the corresponding expression manifold. The high-intensity and low-intensity samples in the subspace form a training matrix, denoted as $\mathbf{T}_\xi$, as follows:
$$\mathbf{T}_\xi = [\mathbf{Y}^1 \quad \mathbf{Y}^\xi] = [\mathbf{A}^T\mathbf{X}^1 \quad \mathbf{A}^T\mathbf{X}^\xi], \tag{34}$$
where $\xi$ ($0 \leq \xi \leq 1$) represents the intensity of the low-intensity images. Figures 3(b) and 3(c) demonstrate the training data $\mathbf{T}_\xi$ with two different values of $\xi$ on the CK+ database.

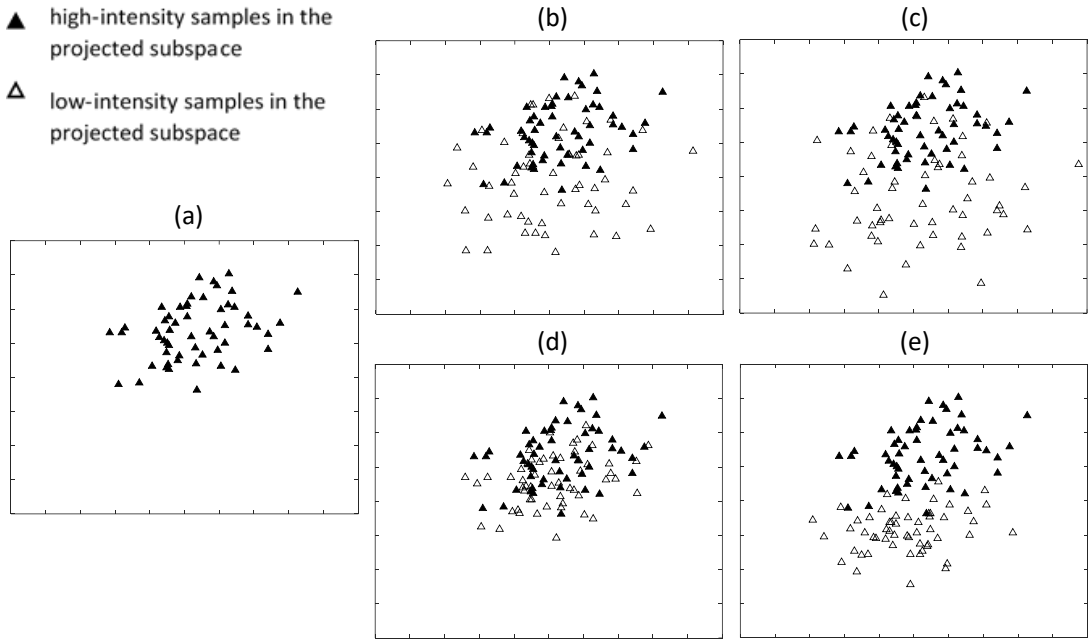

Figure 3. The representation of the feature vectors (FV) of happiness (HA) on the CK+ database, after SLPM: (a) HA, i.e. high-intensity expression samples are applied to SLPM, (b) HA + low intensity FV with $\xi = 0.9$, (c) HA + low intensity FV with $\xi = 0.7$, (d) HA + generated FV with $\theta_{ne} = 0.9$, and (e) HA + generated FV with $\theta_{ne} = 0.7$.

Conventional manifold-learning methods map training samples, irrespective of how strong the expressing images are, as close as possible after transformation. This results in limited performance in terms of generalization. In our feature-generation algorithm, the subspace learning method, SLPM, is first applied to features extracted from high-intensity expressions. Then, features extracted from low-intensity expressions are mapped to the learned subspace. As observed in Figure 4, features extracted from low-intensity expressions are located farther from the core samples (formed by high-intensity expressions) and near the boundary of the manifolds after the mapping.

The high-intensity samples are used to determine the centroid of an expression manifold, while those low-intensity samples are for representing the manifold boundary. The feature vectors are multi-dimensional, so a large number of low-intensity samples are required to represent the manifold boundary faithfully. However, only a small number of low-intensity images are available from the training video sequences. Furthermore, most of the existing expression databases have static images only. To solve this problem, we propose generating more low-intensity feature vectors for each expression class, so that the manifold learned for each expression class will be more accurate. In this paper, we consider the recognition of six facial expressions, i.e. anger, disgust, fear, happiness, sadness, and surprise. In addition to these facial expressions, we also consider the neutral expression in the proposed feature-generation method.

Let $\{\boldsymbol{x}^0_{s_1}, \boldsymbol{x}^0_{s_2}, \ldots, \boldsymbol{x}^0_{s_p}\} \in \mathbb{R}^D$ be the set of feature vectors extracted from neutral face images, where $\boldsymbol{x}^0_{s_i}$ is the feature vector of the neutral face image belonging to the subject $s_i$, and $p$ is the number of the subjects in the dataset. The expression images of the subject $s_i$ are denoted as
$$\mathbf{X}^1_{s_i} = [\boldsymbol{x}^1_{s_{i,1}}, \boldsymbol{x}^1_{s_{i,2}}, \ldots, \boldsymbol{x}^1_{s_{i,r}}], \tag{35}$$
where $r$ is the number of expression images belonging to $s_i$ and $\boldsymbol{x}^1_{s_{i,j}}$ is the feature vector extracted from the $j$th expression image of $s_i$. Then, the feature matrix for all the expressions is formed as follows:
$$\mathbf{X}^1_s = [\mathbf{X}^1_{s_1}, \mathbf{X}^1_{s_2}, \ldots, \mathbf{X}^1_{s_p}]. \tag{36}$$

Figure 4. The subspace learned using SLPM, with local descriptors "LPQ", based on the dataset named CK+: (a) the mapped features extracted from high-intensity expression images and neutral face images, (b) the mapped features extracted from high-intensity and low-intensity ($\xi = 0.7$) images, and (c) the mapped features extracted from high-intensity and low-intensity ($\xi = \{0.9, 0.8, 0.7, 0.6, 0.5, 0.4\}$) images.

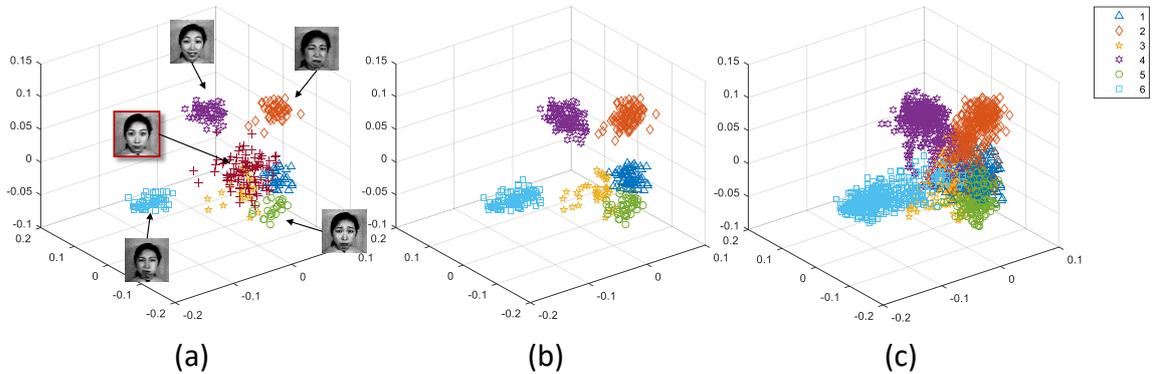

The proposed sample-generation method operates in the learned subspace. Thus, the feature vectors extracted from the neutral face images and the expression images are all mapped to the learned subspace using the projection matrix $\mathbf{A}$ learned from $\mathbf{X}_s^1$, as follows:

$$\mathbf{Y}^1 = \mathbf{A}^T\mathbf{X}_s^1 = \left[\mathbf{Y}_{s_1}^1, \mathbf{Y}_{s_2}^1, \dots, \mathbf{Y}_{s_p}^1\right], \text{ and} \tag{37}$$

$$\mathbf{Y}^0 = \mathbf{A}^T\mathbf{X}_s^0 = \left[\mathbf{y}_{s_1}^0, \mathbf{y}_{s_2}^0, \dots, \mathbf{y}_{s_p}^0\right]. \tag{38}$$

Equations (37) and (38) represent the set of feature vectors of high-intensity expressions and neutral expressions of all subjects, respectively, in the subspace.

The proposed feature-generation method generates low-intensity feature vectors based on vector-pairs selected from two different sets: (1) vector-pairs from $\mathbf{Y}_{s_i}^1$, (2) vector-pairs from $\mathbf{Y}_{s_i}^1$ and $\mathbf{y}_{s_i}^0$. In the following sections, we will describe the feature-generation method with respect to two different vector-pairs.

4.2.2. *Vector-pairs from $\mathbf{Y}_{s_i}^1$ and $\mathbf{y}_{s_i}^0$*

Let $\overline{\mathbf{Y}}_{s_i}^{\theta_{ne}} = \left[\mathbf{y}_{s_{i,1}\to 0}^{\theta_{ne}}, \mathbf{y}_{s_{i,2}\to 0}^{\theta_{ne}}, \dots, \mathbf{y}_{s_{i,r}\to 0}^{\theta_{ne}}\right]$ be the feature matrix of possible low-intensity expressions with an intensity of $\theta_{ne}$ ($0 < \theta_{ne} < 1$) belonging to the subject $s_i$, where $\mathbf{y}_{s_{i,j}\to 0}^{\theta_{ne}}$ is the corresponding low-intensity feature vector generated using $\mathbf{y}_{s_{i,j}}^1$ and $\mathbf{y}_{s_i}^0$. In the rest of the paper, the arrow "→" indicates the direction of the feature vectors to be generated, with 0 and 1 being a neutral face image and a face image with the highest intensity, respectively. $\mathbf{y}_{s_{i,j}\to 0}^{\theta_{ne}}$ means that the feature vector is generated in the direction from $\mathbf{y}_{s_{i,j}}^1$ to $\mathbf{y}_{s_i}^0$ where $\mathbf{y}_{s_{i,j}}^1$ is the mapped feature vector extracted from the *j*th expression image of $s_i$.

Figure 5. The representation of the sample-generation process based on (a) feature vectors extracted from high-intensity images and neutral-face images, and (b) feature vectors extracted from high-intensity images.

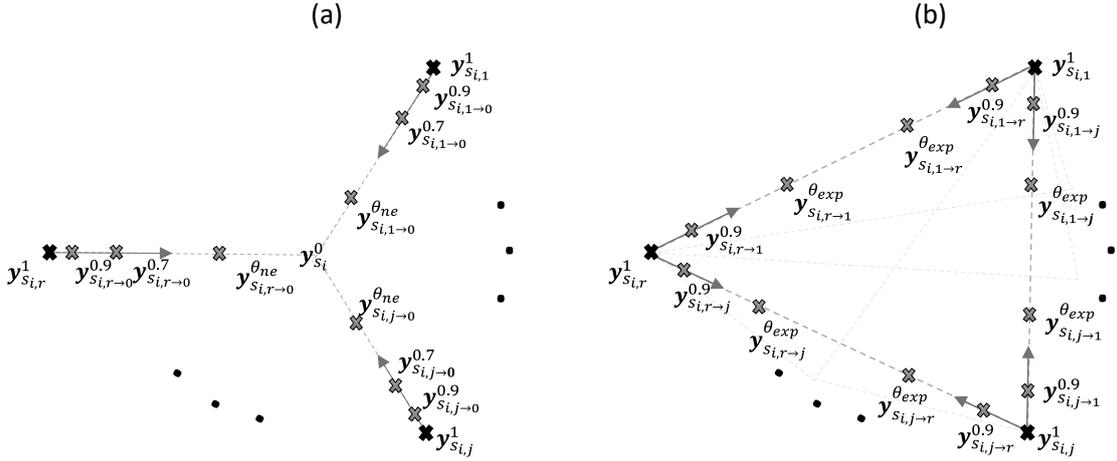

A set of feature vectors extracted from an expression video sequence, which starts from a neutral-expression face to the highest intensity of an expression, can be perceived as a path from the reference center, i.e. the neutral manifold, to a particular expression manifold wherein the distance of an expression manifold from the center is directly proportional to the intensity of the expression [36]. Therefore, for databases consisting of only static expression images, the feature matrix of possible low-intensity expressions can be obtained by assuming that the relation between the distance from $\mathbf{y}_{s_{i,j}\to 0}^{\theta_{ne}}$ to

$y^0_{s_i}$ and the expression intensity is linear. As illustrated in Fig. 5(a), the low-intensity feature vector $y^{\theta_{ne}}_{s_{i,j}\to 0}$, belonging to $s_i$, can be computed as follows:

$$y^{\theta_{ne}}_{s_{i,j}\to 0} = \theta_{ne} \cdot y^1_{s_{i,j}} + (1-\theta_{ne}) \cdot y^0_{s_i}. \tag{39}$$

Figs 3(d) and 3(e) outline the training data with the feature generation using neutral images when $\theta_{ne} = 0.9$ and $\theta_{ne} = 0.7$, respectively. As seen in Fig. 3, both the absolute low-intensity feature vectors and the possible low-intensity feature vectors generated by linear interpolation have a similar structure.

### 4.2.3. Vector-pairs from $Y^1_{s_i}$

The respective expression manifolds can be far from each other in the learned subspace. For this reason, more features between expression manifolds are also needed. In the previous section, we proposed the idea that the feature vectors extracted from low-intensity expression images should be distant from the corresponding manifold centre, thus, this can enhance the generalizability of the learned manifold. Using a similar idea, more features that are distant from the manifold centres can be generated using vector-pairs from the feature matrix of high-intensity expressions of the same subject, $Y^1_{s_i}$, as illustrated in Fig. 5(b). A feature vector $y^{\theta_{exp}}_{s_{i,j}\to k}$, which lies on the line from the $j$th expression-vector of $s_i$, $y^1_{s_{i,j}}$, to the $k$th expression-vector of $s_i$, $y^1_{s_{i,k}}$, with a weight $\theta_{exp}$ ($0 < \theta_{exp} < 1$) can be computed as follows:

$$y^{\theta_{exp}}_{s_{i,j}\to k} = \theta_{exp} \cdot y^1_{s_{i,j}} + (1-\theta_{exp}) \cdot y^1_{s_{i,k}}, \tag{40}$$

Suppose that $c_j = l(y^1_{s_{i,j}})$ and $c_k = l(y^1_{s_{i,k}})$ are the expression classes of the $j$th and the $k$th expression vectors, respectively, and $n_{i,c_j}$ and $n_{i,c_k}$ are the number of expression-vectors of expression classes $c_j$ and $c_k$, respectively, belonging to subject $s_i$. Then, a total of $n_{i,c_j} n_{i,c_k}$ feature vectors can be generated. The feature matrix consisting of the generated features using the pairs from $Y^1_{s_i}$ can be denoted as follows:

$$\overline{Y}^{\theta_{exp}}_{s_i,exp} = \left[ y^{\theta_{exp}}_{s_{i,1}\to 2}, y^{\theta_{exp}}_{s_{i,1}\to 3}, \ldots, y^{\theta_{exp}}_{s_{i,1}\to r}, \ldots, y^{\theta_{exp}}_{s_{i,r}\to 1}, y^{\theta_{exp}}_{s_{i,r}\to 2}, \ldots, y^{\theta_{exp}}_{s_{i,r-1}\to r} \right]. \tag{41}$$

The training matrix, $T_\theta$, is updated to $\overline{T}_\theta$, which is used as a static database, as follows:

$$\overline{T}_\theta = \left[ Y^1 \quad \overline{Y}^{\theta_{ne}}_{ne} \quad \overline{Y}^{\theta_{exp}}_{exp} \right], \tag{42}$$

where $\overline{Y}^{\theta_{ne}}_{ne} = \left[ \overline{Y}^{\theta_{ne}}_{s_1,ne}, \overline{Y}^{\theta_{ne}}_{s_2,ne}, \ldots, \overline{Y}^{\theta_{ne}}_{s_p,ne} \right]$ and $\overline{Y}^{\theta_{exp}}_{exp} = \left[ \overline{Y}^{\theta_{exp}}_{s_1,exp}, \overline{Y}^{\theta_{exp}}_{s_2,exp}, \ldots, \overline{Y}^{\theta_{exp}}_{s_p,exp} \right]$.

In our experiments, we vary the $\theta_{ne}$ and the $\theta_{exp}$ values from 0.7 to 0.9. Figure 2 lists the overall flow of the proposed algorithm.

When a feature vector is generated, it is checked whether or not it is closest to its corresponding manifold class. Furthermore, the feature vectors are generated solely for the pairs of clusters that are in close proximity to each other in the learned subspace.

Table 3. A comparison of the number of images for different expression classes in the databases used in our experiments

|           | BAUM-2 | CK+ | JAFFE | TFEID |
|-----------|--------|-----|-------|-------|
| **Anger**     | 80  | 45  | 30  | 34  |
| **Disgust**   | 32  | 59  | 29  | 40  |
| **Fear**      | 35  | 25  | 32  | 40  |
| **Happiness** | 139 | 69  | 31  | 40  |
| **Sadness**   | 68  | 28  | 31  | 39  |
| **Surprise**  | 83  | 82  | 30  | 36  |
| **Neutral**   | 99  | 106 | 30  | 39  |
| **TOTAL**     | **536** | **414** | **213** | **268** |

## 5. Experimental set-up and results

### 5.1. Experimental setup

In our experiments, four facial-expression databases were used to show the robustness and performances of the proposed methods: 1) Bahcesehir University Multilingual Affective Face Database (BAUM-2) [37], 2) Extended Cohn-Kanade (CK+) [38] database, 3) Japanese Female Facial Expression (JAFFE) [39] database, and 4) Taiwanese Facial Expression Image Database (TFEID) [40].

The BAUM-2 multilingual database consists of short videos extracted from movies. In our experiments, an image dataset, namely BAUM-2i, consisting of images with peak expressions from the videos in BAUM-2, is considered. There are 829 face images from 250 subjects in the BAUM-2i static expression dataset, which express 6 basic emotions. However, only 536 of them, which have their facial-feature points provided, are considered in our experiments. Since the BAUM-2 database was created by extracting images from movies, the images are close to real-life conditions (i.e. under pose, age, and illumination variations, etc.), and are more challenging than those in acted databases.

The CK+ dataset contains a total of 593 posed sequences across 123 subjects, of which 304 of the sequences have been labelled with one of the six discrete emotions, which are anger, disgust, fear, happiness, sadness, and surprise. Each sequence starts with a neutral face and ends with a frame of peak expression. The last frame of each sequence, and the first frames of the sequences that have unique subject labels, as well as their landmarks provided, are used for recognition. There are a total of 414 face images. Note that some of the first frames are also discarded because the expressed emotions are of low intensity. JAFFE consists of 213 images from 10 Japanese females, which express 6 basic emotions – anger, disgust, fear, happiness, sadness, surprise – and neutral. JAFFE is also a widely used acted database, which means that it was recorded in a controlled environment. The TFEID database contains 268 images, with the six basic expressions and the neutral expression, from 40 Taiwanese subjects. Like CK+ and JAFFE, TFEID is also an acted database.

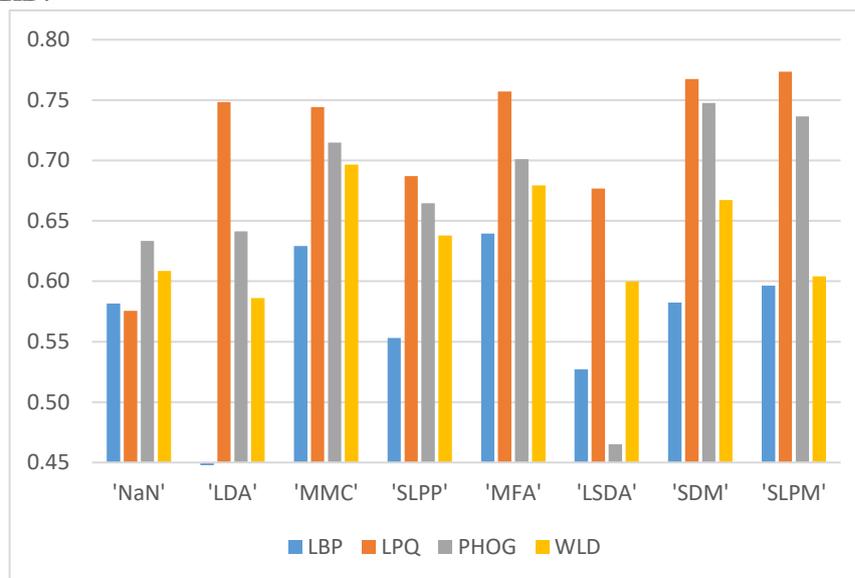

Figure 6. The recognition rates of the different subspace methods, with different local descriptors, based on a combined dataset of BAUM-2, CK+, JAFFE & TFEID.

Each of the above-mentioned databases has its own characteristics. Table 3 shows the number of images for each expression class for the different databases. Although some of the databases also have

Figure 7. Recognition rates of our proposed method in terms of different dimensions.

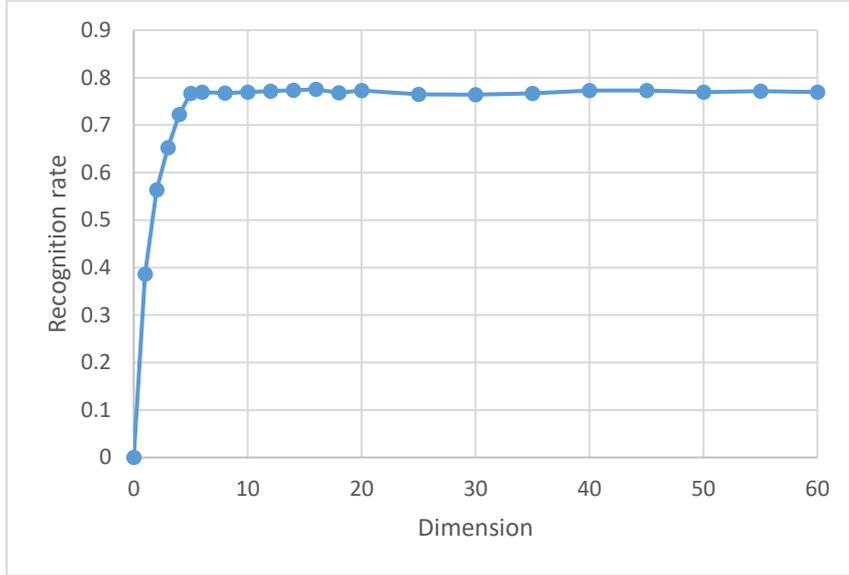

the contempt expression, only the six basic prototypical facial expressions (i.e. anger, disgust, fear, happiness, sadness, and surprise), as well as the neutral facial expression, are considered in our experiments. Please note that neutral facial expression has been used only for creating feature vectors

Table 4. The comparison of recognition rates obtained by using low-intensity images with different $l$ values on the CK+ database, using the LPQ feature.

| METHODS | CK+ |
|---|---|
| **SLPM** | 94.81% |
| **SLPM + $\xi = 0.9$** | **95.45%** |
| **SLPM + $\xi = 0.8$** | 94.16% |
| **SLPM + $\xi = 0.7$** | 93.51% |
| **SLPM + $\xi = 0.6$** | 91.88% |

of low-intensity expressions.

Subspace-learning methods are often applied to feature vectors formed by the pixel intensities of face images. In our method, features are first extracted using the state-of-the-art local descriptors, and then a subspace-learning method is applied for manifold learning and dimensionality reduction. The usual way of using local descriptors is to divide a face image into a number of overlapping or non-overlapping regions, then extract features from these regions, and finally concatenate them to form a single feature vector. In this way, local information, as well as spatial information, can be obtained. Another way of using local descriptors is to consider only the regions that have more salient information about the considered expression classes. Following this idea, features extracted from the eye and mouth regions are used in [41], which showed that features extracted from these regions only can achieve higher recognition rates than those extracted by dividing face images into sub-regions.

In our experiments, face images from the different databases are all scaled to the size of 126×189 pixels, with a distance of 64 pixels between the two eyes. To determine the eye and mouth windows, the facial landmarks, i.e. the eyes and mouth corners, are used. If facial landmarks are not provided for a database, the required facial-feature points are marked manually. The eye window and the mouth window are further divided into 12 and 8 sub-regions, respectively. The nearest neighbour classifier

Table 5. The comparison of subspace learning methods on different datasets, with the LPQ descriptor being used with nearest neighbor classifier.

|  | BAUM-2 | CK+ | JAFFE | TFEID |
|---|---|---|---|---|
| **MFA** | 62.01% | 93.83% | 89.07% | 91.70% |
| **SDM** | 62.01% | 93.51% | 89.07% | 92.58% |
| **SLPM** | 62.93% | 94.81% | 90.71% | 93.45% |
| **SLPM** + $\theta_{exp} = 0.9 + \theta_{ne} = 0.9$ | **63.62%** | 94.81% | 91.26% | 93.45% |
| **SLPM** + $\theta_{exp} = 0.8 + \theta_{ne} = 0.8$ | 62.93% | **96.10%** | 91.26% | **94.32%** |
| **SLPM** + $\theta_{exp} = 0.7 + \theta_{ne} = 0.7$ | 63.16% | 95.13% | **91.80%** | 93.89% |
| **SLPM** + $\theta_{exp} = 0.6 + \theta_{ne} = 0.6$ | 62.01% | 94.16% | 90.71% | 93.45% |

and SVM with linear kernel are used in the experiments.

### 5.2. Experimental results

In this section, we evaluate the performances of our proposed method, using four different descriptors, on the four different databases. We also compare our method with four subspace-learning methods, as well as without using any subspace-learning method.

Firstly, the four acted databases, i.e. BAUM-2, CK+, JAFFE, and TFEID, are combined to form a single dataset, called COMB4, so that we can better measure the general performances of the different subspace-learning methods and the descriptors.

Figure 6 shows that MFA, SDM, and SLPM are the three best subspace-learning methods, which outperform the other subspace-learning methods. The LPQ local descriptor achieves the highest recognition rates, for the different subspace-learning methods, on COMB4. Therefore, the subspace-learning methods, MFA and SDM, and the local descriptor, LPQ, are chosen to further compare the performance of the proposed method on each of the individual datasets. In Figure 6, we can also observe that SDM outperforms most of the subspace-learning methods, except SLPM, because the intra-class spread is adjustable. Furthermore, SDM is also computationally simpler than the other compared methods, but it does not incorporate the local geometry of the data. In our proposed method, information about local structure is incorporated into the objective function. Thus, SLPM can achieve higher recognition rates than SDM.

Table 6. Comparison of the runtimes (in milliseconds) required by the different subspace learning methods (MFA, SDM, and SLPM) on different datasets, with the LPQ descriptor used.

|  | BAUM-2 | CK+ | JAFFE | TFEID |
|---|---|---|---|---|
| **MFA** | 96 | 69 | 45 | 51 |
| **SDM** | 151 | 133 | 120 | 118 |
| **SLPM** | **65** | **37** | **23** | **25** |

Figure 7 shows the recognition rates of SLPM on COMB4, with the dimensionality of the subspace varied. The results show that SLPM has converged to its highest recognition rate, when the dimensionality is lower than 10. In other words, our method is still very effective even at a low dimensionality. Based on these results, we set the subspace dimensionality at 11 in the rest of the experiments.

To investigate the effect of the use of images of expression with low intensities, several experiments have been conducted on the CK+ database. As shown in Table 4, the recognition rate is the highest when $\xi = 0.9$. Table 5 and Table 6 show the recognition rates of the three subspace-learning methods, MFA, SDM and SLPM, as well as SLPM, using feature generation with different $\theta_{exp}$ and $\theta_{ne}$ values, with the LPQ descriptor, on the four different databases using nearest neighbour classifier and SVM classifier, respectively. It can be found that SLPM achieves the best classification performance again, when compared to the other methods. The classification performance is further improved by up to 2%, when feature generation is employed. Furthermore, as observed in Tables 5 and 6, the nearest neighbour classifier outperforms the SVM classifier in most of the databases. Lastly, additional experiments were conducted to validate the efficiency of the proposed subspace learning methods. Table 7 tabulates the runtimes in milliseconds for each of the subspace learning methods. We can see that SLPM is twice as fast as MFA, which solves the generalized eigenvalue problem instead of calculating eigenvalue decomposition like SLPM. SDM is much slower than MFA and SLPM.

## 6. Conclusion

In this paper, we have proposed a subspace-learning method, named Soft Locality Preserving Map (SLPM), which uses the neighbourhood and class information to construct a projection matrix for mapping high-dimensional data to a meaningful low-dimensional subspace. The difference between the within-class and between-class matrices is used to define the objective function, rather than the Fisher criteria, in order to avoid the singularity problem. Also, a parameter *β* is added to control the within-class spread, so that the overfitting problem can be solved. The robustness and the generalizability of SLPM have been analysed on four different databases, using four different state-of-the-art descriptors and two different classifiers, and SLPM has been compared with other subspace-learning methods. Moreover, we have proposed using low-intensity expression images to learn a better manifold for each expression class. By taking advantage of domain-specific knowledge, we have proposed two methods of generating new low-intensity features in the subspace. Our experiment results have shown that SLPM outperforms the other subspace-learning methods, and is a good alternative to performing dimensionality reduction on high-dimensional datasets. Our experiment results, also, have shown that the proposed feature-generation method can further increase the recognition rates.

Table 7. The comparison of subspace learning methods on different datasets, with the LPQ descriptor being used with SVM classifier.

|  | BAUM-2 | CK+ | JAFFE | TFEID |
|---|---|---|---|---|
| **MFA** | 61.10% | 92.21% | **91.26%** | 91.70% |
| **SDM** | 60.18% | 92.21% | 89.62% | 92.58% |
| **SLPM** | 63.16% | 92.53% | 89.62% | 93.01% |
| **SLPM + $\theta_{exp} = 0.9 + \theta_{ne} = 0.9$** | **63.84%** | 92.86% | **91.26%** | 94.76% |
| **SLPM + $\theta_{exp} = 0.8 + \theta_{ne} = 0.8$** | 62.47% | 93.83% | **91.26%** | **95.20%** |
| **SLPM + $\theta_{exp} = 0.7 + \theta_{ne} = 0.7$** | 62.24% | **94.48%** | 89.07% | 94.32% |
| **SLPM + $\theta_{exp} = 0.6 + \theta_{ne} = 0.6$** | 61.56% | **94.48%** | 88.52% | 94.32% |